\RequirePackage{amsthm}
\documentclass[sn-mathphys-num]{sn-jnl}% Math and Physical Sciences Numbered Reference Style 
\usepackage{graphicx}%
\usepackage{multirow}%
\usepackage{amsmath,amssymb,amsfonts}%
\usepackage{amsthm}%
\usepackage{mathrsfs}%
\usepackage[title]{appendix}%
\usepackage{xcolor}%
\usepackage{textcomp}%
\usepackage{manyfoot}%
\usepackage{booktabs}%
\usepackage{algorithm}%
\usepackage{algorithmicx}%
\usepackage{algpseudocode}%
\usepackage{listings}%
\usepackage{tikz}
\usepackage{pgfplots}
\usepackage{multirow} 
\usepackage{multicol}
\usepackage{tcolorbox}
\usepackage{threeparttable}
\pgfplotsset{compat=1.11,
        /pgfplots/ybar legend/.style={
        /pgfplots/legend image code/.code={%
        \draw[##1,/tikz/.cd,bar width=3pt,yshift=-0.2em,bar shift=0pt]
                plot coordinates {(0cm,0.8em)};},
},
}
\newcommand\revise[1]{{\color{black}#1}}
\theoremstyle{thmstyleone}%
\theoremstyle{thmstyletwo}%

\theoremstyle{thmstylethree}%

\begin{document}

\title{Unveiling Scoring Processes: Dissecting the Differences  between LLMs and Human Graders in Automatic Scoring}


% \author{\fnm{Xuansheng} \sur{Wu}}\email{xuansheng.wu@uga.edu}
% \author{\fnm{Padmaja Pravin} \sur{Saraf}}\email{padmajapravin.saraf@uga.edu}
% \author{\fnm{Gyeong-Geon} \sur{Lee}}\email{ggleeinga@uga.edu}
% \author{\fnm{Ehsan} \sur{Latif}}\email{ehsan.latif@uga.edu}
% \author{\fnm{Ninghao} \sur{Liu}}\email{ninghao.liu@uga.edu}
% \author{\fnm{Xiaoming} \sur{Zhai}}\email{xiaoming.zhai@uga.edu}

% \affil{\orgname{University of Georgia}, \orgaddress{\street{Herty Drive}, \city{Athens}, \postcode{30602}, \state{Georgia}, \country{USA}}}

% %%=============================================================%%
% %% GivenName	-> \fnm{Joergen W.}
% %% Particle	-> \spfx{van der} -> surname prefix
% %% FamilyName	-> \sur{Ploeg}
% %% Suffix	-> \sfx{IV}
% %% \author*[1,2]{\fnm{Joergen W.} \spfx{van der} \sur{Ploeg} 
% %%  \sfx{IV}}\email{iauthor@gmail.com}
% %%=============================================================%%

\author[1]{\fnm{Xuansheng} \sur{Wu}}\email{xuansheng.wu@uga.edu}

\author[1]{\fnm{Padmaja Pravin} \sur{Saraf}}\email{padmajapravin.saraf@uga.edu}

\author[2]{\fnm{Gyeonggeon} \sur{Lee}}\email{gyeonggeon.lee@nie.edu.sg}

\author[3,4]{\fnm{Ehsan} \sur{Latif}}\email{ehsan.latif@uga.edu}

\author[1]{\fnm{Ninghao} \sur{Liu}}\email{ninghao.liu@uga.edu}

\author*[3,4]{\fnm{Xiaoming} \sur{Zhai}}\email{xiaoming.zhai@uga.edu}

\affil[1]{\orgdiv{School of Computing}, \orgname{University of Georgia}, \orgaddress{\street{D. W. Brooks Drive}, \city{Athens}, \postcode{30602}, \state{Georgia}, \country{United States}}}

\affil[2]{\orgdiv{National Institute of Education}, \orgname{Nanyang Technological University}, \orgaddress{\street{1 Nanyang Walk}, \city{Singapore}, \postcode{637616}, \country{Singapore}}}

\affil[3]{\orgdiv{AI4STEM Education Center}, \orgname{University of Georgia}, \orgaddress{\street{110 Carlton St}, \city{Athens}, \postcode{30602}, \state{Georgia}, \country{United States}}}
\affil[4]{\orgdiv{Department of Mathematics, Science, and Social Studies Education}, \orgname{University of Georgia}, \orgaddress{\street{110 Carlton St}, \city{Athens}, \postcode{30602}, \state{Georgia}, \country{United States}}}


\abstract{Large language models (LLMs) have demonstrated strong potential in performing automatic scoring for constructed response assessments. While constructed responses graded by humans are usually based on given grading rubrics, the methods by which LLMs assign scores remain largely unclear. It is also uncertain how closely AI's scoring process mirrors that of humans or if it adheres to the same grading criteria. To address this gap, this paper uncovers the grading rubrics that LLMs used to score students' written responses to science tasks and their alignment with human scores. We also examine whether enhancing the alignments can improve scoring accuracy. Specifically, we prompt LLMs to generate analytic rubrics that they use to assign scores and study the alignment gap with human grading rubrics. Based on a series of experiments with various configurations of LLM settings, we reveal a notable alignment gap between human and LLM graders. While LLMs can adapt quickly to scoring tasks, they often resort to shortcuts, bypassing deeper logical reasoning expected in human grading. We found that incorporating high-quality analytical rubrics designed to reflect human grading logic can mitigate this gap and enhance LLMs' scoring accuracy. \revise{These results underscore the need for a nuanced approach when applying LLMs in science education and highlight the importance of aligning LLM outputs with human expectations to ensure efficient and accurate automatic scoring.}
}




\maketitle

\section{Introduction}
In science education, educators develop scientific items to evaluate student abilities in logical thinking of understanding scientific phenomena~\cite{dataset,wilson2024using}. 
As automatic scoring helps students receive feedback to their responses in a timely fashion, it becomes a critical step in today's science education~\cite{automatic-guidance,guo2024using}. 
Inspired by the substantial progress in the field of natural language processing, some researchers~\cite{MeNSP,CoT,zhai2023chatgpt} propose to develop automatic scoring systems based on advanced large language models (LLMs). 
\revise{LLMs~\cite{Gemini,Llama,jiang2024mixtral} are machine learning models trained to understand and generate human natural language. Modern LLMs are trained with a massive corpus from the Internet, and they can help humans with various daily tasks, such as writing emails, planning travel, and so on. 
Different from conventional machine learning models, these LLMs have demonstrated a strong ability to quickly adapt to new downstream tasks (i.e., automatic scoring in this paper) with limited or even without any training samples~\cite{in-context}. 
Therefore, these advanced LLMs are leveraged to assist human educators for various education purposes, such as tutoring~\cite{parikh2019looking,graesser2005autotutor}, assessment and evaluations~\cite{MeNSP,CoT}, and peer-to-peer learning~\cite{zhai2023chatgpt}.  
}
This advancement significantly saves the cost of developing automatic scoring models and thus helps more students benefit from high-quality science education.

However, it is still unclear whether LLM graders behave as human graders, raising the potential risks of applying them to border cases. 
\revise{Practically, if an LLM grader incorrectly assigns a poorer score to a student, it will hurt not only the student's interest in studying but also the teacher's confidence in using these systems. Also, if it incorrectly provides a higher score to a student, it could fail to provide prompt feedback to students and guide their further studying.}
To fill this gap, this study focuses on two research questions: (1) Is there an alignment gap between the human and LLM graders during the scoring reasoning steps? (2) If so, could filling this gap increase LLM graders' scoring accuracy? 
While limited studies have answer these two questions, some attempt to improve the scoring performance of LLM graders. Although some prompt engineering strategies, such as Chain-of-Thoughts~\cite{Chain-of-thought,kojima2022large} and In-context Examples~\cite{in-context}, improve the scoring accuracy of LLM-based systems, most of these LLMs cannot provide as accurate scoring feedback as humans do. 
It is crucial to recognize that these studies find the gap by simply comparing the LLM-generated and human-provided categorical grades without examining the real inherent process of "thoughts" of LLM graders during automatic scoring. 
That is, even if LLMs can provide grading that exactly match human graders, the gap may still exist because LLMs may draw the same conclusion on a path different from humans. 
These unexpected paths usually are referred to \textit{shortcuts}~\cite{shortcuts,tang2023large}, indicating that LLMs find a superficial way to make predictions without following a reasonable logic chain as human experts.

To address these challenges, this study incorporates human-designed analytic rubrics to examine the behavior of LLM graders more inherently. Human graders typically employ a set of detailed rules, called \textit{analytic rubrics}, to evaluate the quality of student responses, where each rule usually reflects a specific required proficiency of students expected by the human educator. The agreement of a student's response with more of these rules generally correlates with higher scores. These rubrics effectively outline the logical path a grader follows in evaluating a student's response. Therefore, rather than simply comparing scores, our research examines the differences between the analytic rubrics generated by LLMs and those crafted by humans, allowing us to explore the inherent differences between LLM and human graders.

\begin{figure*}[h!t]
\caption{\revise{One exemplar assessment item from the datasets\cite{harris2024creating}. We present their human-crafted meta-information, including the task description, holistic rubrics, and analytic rubrics.}}
\label{dataset_example}
\begin{tcolorbox}[colback=gray!5!white,width=\textwidth,colframe=blue!75!black,]
\textbf{Task} \\
This task is measuring a student's proficiency in the following: Develop a model that 
explains how particle motion changes when thermal energy is transferred to or from a 
substance without changing state. Shwan had 3 dishes of water at room temperature. She 
cooled one dish, causing thermal energy to transfer from that dish to surroundings. She 
kept the middle dish at room temperature. She transferred thermal energy into the third 
dish by heating it. Then Shwan dropped a red-coated chocolate candy into each dish. 
Construct a model that shows what is happening to water particles and red dye particles 
in each dish. \\
\\
\textbf{Holistic Rubrics}\\
- Proficient Level: Student develops a model that fully identifies both water and dye particles and their motions while describing that water molecules move faster at higher temperatures (and vice versa).\\
- Developing Level: Student develops a model that partially identifies both water and dye particles and their motion while describing that water molecules move faster at higher temperatures (and vice versa).\\
- Beginning Level: Student does not at all develop a model that identifies both water and dye particles and their motion while describing that water molecules move faster at higher temperatures (Vice Versa).\\
\\
\textbf{Analytic Rubrics}\\
- Rule 1: Water and dye molecules move slowly when the water is cold, faster when at room temperature, and faster when the water is hot.\\
- Rule 2: The key identifies water and dye particles.\\
- Rule 3: The key identifies the particle’s motion (faster/slower).\\
- Rule 4: The answer is a meaningful sentence.
\end{tcolorbox}
\end{figure*}

Our experiments span 12 different assessment items in science education, majorly covering topics in physics. Initial findings reveal notable differences between the LLM-generated rubrics and those crafted by human graders, indicating a significant alignment gap. Specifically, we observe that providing graded responses as examples cannot fill this gap while introducing holistic rubrics could help. Surprisingly, our further qualitative analysis suggests that providing LLMs with graded student responses does not inherently teach them to understand the question better; rather, it encourages LLM graders to use shortcuts in scoring. This observation against traditional recognition in automatic scoring that providing in-context graded responses could help LLM graders make more accurate grades. These findings caution against overly simplistic applications of LLMs in automatic scoring. Finally, our experiment demonstrates that integrating high-quality analytic rubrics can improve the performance of LLM graders in automatic scoring, highlighting the importance of addressing the misalignment gap to enhance their effectiveness.

\section{Related Works}
\revise{Research on leveraging LLMs for automatic scoring is growing rapidly as this new perspective could provide consistent, less-bias, and personalized grading feedback for students compared with traditional grading methods~\cite{yan2024practical,haque2022semantic,han2023fabric,latif2023artificial}.} 
Early work focuses on fine-tuning LLMs for automatic scoring~\cite{latif2024fine,organisciak2023beyond}, which typically provides accurate grading labels as human graders. Specifically, they may collect thousands of labeled student responses or a large corpus related to a specific assessment item for model training\cite{zhai2024ai}. While this procedure helps LLMs to adapt to the assessment items and provide high-quality feedback, it also raises a significant challenge in preparing such training data, which is costly and time-consuming. 
Other researchers~\cite{MeNSP} switch to explore whether it is possible to transform the automatic scoring task into one of the LLMs' pre-training tasks. In this way, the development of an LLM-based automatic scoring system does not require training.
Although this approach saves costs from both collecting datasets and training models, it also remains two critical issues. One is that such systems cannot provide highly accurate scoring labels since the LLMs have no chance to update their parameters to adapt to the automatic scoring task. 
Also, it leads to an ethical issue of whether those un-trained LLMs are reliable enough to serve our interested scoring tasks.
Recently, researchers~\cite{CoT,bewersdorff2023assessing,wu2022automatic,cohn2024chain,xia2024empirical} have shown that the first issue could be partially solved by using more advanced LLMs or designing dedicated prompt-engineering strategies. 
In particular, these studies indicate that providing more graded examples within the prompting templates or prompting LLMs to provide explanations along with their predicted scoring results could enhance LLMs to provide more accurate scoring results. 
However, their results reveal that those enhanced LLM graders could mimic human graders for those simpler assessment items, while they may fail to provide human-like feedback on those harder ones. 
More importantly, the second issue about ethical concerns of this prompting-based approach is still unclear. 

Since automatic scoring is a sensitive task directly connected with the assessment and evaluation of un-adult students, it is crucial to know on what basis LLM performs the scoring tasks and how it concludes the final score. To achieve validity in the automatic scoring task, it becomes vital to have explainability in this automatic scoring procedure~\cite{zhao2024explainability}. 
 \revise{
Asking LLMs to provide a chain-of-thought explanation for their predictions is a practical way to explore their internal decision-making procedures~\cite{wei2022chain,wu2024usable}. By leveraging this technique, researchers~\cite{CoT,cohn2024chain} effectively collect the explanation while grading each student response, and the human educators evaluate the reliability of each individual feedback based on these explanations. However, this approach can not provide a full picture of the internal understanding of LLMs at an item level.
To address this research gap, this study proposes to provide the explainability to the automatic scoring task by prompting LLMs to provide the rubrics of each assessment item. By further analyzing these LLM-generated rubrics, human educators can build a comprehensive recognition of the abilities of LLM graders over item levels.}

\section{Methodology}
\subsection{Datasets}
This study experiments on a dataset by asking middle school students to describe scientific models accounting for science phenomena~\cite{harris2024creating,zhai2022applying}. The dataset includes 7 assessment tasks, out of which two assessment tasks have binomial scoring rubrics, and the remaining five tasks have trinomial scoring rubrics. \revise{We provide one example of these assessment items in Figure~\ref{dataset_example}.} 
Each assessment task could include multiple assessment items, and some ask the students to draw a figure to respond. By ignoring these items requiring visual responses, we finally selected 12 assessment items, each of which contains a question description (background and question), rubrics (analytical and holistic rubrics), and the number of scoring levels (binomial or trinomial). \revise{Please note that human experts craft all the information (i.e., questions description, holistic and analytic rubrics, scoring levels) of the assessment items.} Here, holistic rubrics evaluate the student response as a whole and do not break the assessment down into different specific rules. It considers the overall student response and then grades the response based on the overall quality. \revise{On the other hand, analytic rubrics break down the assessment with multiple rules, where each of these rules of the assessment is evaluated separately.} It provides detailed criteria based on which the assessment is graded. Each of the criteria has its own rating scale. In the end, all the criteria scores are then summed up to come to the total score or the grade of the student response.   
For each assessment task, we collected around 800 student responses~\cite{zhai2022applying}. Each student response is annotated according to the rubrics by humans into different levels, such as ``Beginning'', ``Developing'', or ``Proficient''. We randomly selected 100 labeled student responses from each assessment item with balanced grading levels and used them to test the automatic scoring performance of LLM graders.

\subsection{Experimental Designs} 
\begin{figure*}
\caption{\revise{Basic prompting templates for analytic rubric generation and automatic scoring tasks. Notes: (1) \textit{\%\%} are comment lines for demonstration purposes, which are not real prompts; (2) \textit{System}, \textit{User} and \textit{Agent} are official identifiers from the official templates for those instruction-tuned LLMs; (3) $\{$\textit{xxxx}$\}$ are slots for filling corresponding contents; (4) \_\_xxxx\_\_ refers to boldface and - xxxx refers to bullet points in markdown, which can be recognized by LLMs.}}
\label{fig:templates}
\begin{tcolorbox}[colback=gray!5!white,colframe=blue!75!black,title=Template-1: Generating analytic rubrics with oneshot examples.]
\textit{\%\% Round-1: Instruction on generating analytic rubrics with role-play.}\\
\textbf{System:} The agent is an impartial science educator working in a middle school. His job is working under the supervision of the User.\\\
\textbf{User:} Your job is to provide the analytic rubric for a science item.\\
\textbf{Agent:} I will begin to work on my job.
\\
\\
\textit{\%\% Round-2: Providing in-context examples for analytic rubric generation.}\\
\textbf{User:}\\- \_\_Task:\_\_ 
$\{$Task Description from Example Item 1$\}$\\
- \_\_Total Points:\_\_ 
$\{$Total Grading Levels from Example Item 1$\}$\\
\textbf{Agent:}\\- \_\_Analytic Rubric:\_\_ 
$\{$Human-written Analytic Rubric from Example Item 1$\}$\\
\\
\textit{\%\% Round-3: Requesting analytic rubric from the target item.}\\
\textbf{User:}\\- \_\_Task:\_\_ 
$\{$Task Description from Testing Item$\}$\\
- \_\_Total Points:\_\_ 
$\{$Total Grading Levels from Testing Item$\}$\\
\textbf{Agent:}
\end{tcolorbox}

\begin{tcolorbox}[colback=gray!5!white,colframe=blue!75!black,title=Template-2: Grading student responses with oneshot examples for an assessment item.]
\textit{\%\% Round-1: Instruction on automatic scoring with role-play.}\\
\textbf{System:} The agent is an impartial science educator working in a middle school. His job is working under the supervision of the User.\\\
\textbf{User:} Your job is to evaluate the quality of the response provided by a student to a science item. Begin your evaluation by providing a short explanation. Be as objective as possible. After providing your explanation, you must classify the response on a scale of ``Beginning", ``Developing," and ``Proficient" by strictly following this format: [[rating]], for example: ``Rating: [[Beginning]]''. Refer to the CONTEXT and RUBRIC while rating.\\
\textbf{Agent:} I will begin to work on my job.
\\
\\
\textit{\%\% Round-2: Providing in-context examples for automatic scoring.}\\
\textbf{User:}\\- \_\_Context:\_\_ 
$\{$Task Description from the assessment item.$\}$\\
- \_\_Rubric:\_\_ 
$\{$Holistic Rubric from the assessment item.$\}$\\
- \_\_Student Response:\_\_ 
$\{$One example student response.$\}$\\
\textbf{Agent:}\\$\{$Ground truth score of the example student response.$\}$
\end{tcolorbox}
\end{figure*}
\begin{figure*}
\caption{\revise{Basic prompting templates for analytic rubric generation and automatic scoring tasks. Notes: (1) \textit{\%\%} are comment lines for demonstration purposes, which are not real prompts; (2) \textit{System}, \textit{User} and \textit{Agent} are official identifiers from the official templates for those instruction-tuned LLMs; (3) $\{$\textit{xxxx}$\}$ are slots for filling corresponding contents; (4) \_\_xxxx\_\_ refers to boldface and - xxxx refers to bullet points in markdown, which can be recognized by LLMs. (continued)}}
\label{fig:templates}
\begin{tcolorbox}[colback=gray!5!white,colframe=blue!75!black,title=Template-1: Generating analytic rubrics with oneshot examples.]
\textit{\%\% Round-3: Requesting automatic scoring for the testing student response.}\\
\textbf{User:}\\- \_\_Context:\_\_ 
$\{$Task Description from the assessment item.$\}$\\
- \_\_Rubric:\_\_ 
$\{$Holistic Rubric from the assessment item.$\}$\\
- \_\_Student Response:\_\_ 
$\{$Texting student response.$\}$\\
\textbf{Agent:}
\end{tcolorbox}
\end{figure*}
We conducted controlled experiments to verify our hypotheses. Specifically, we controlled the behaviors of LLMs by designing different prompts to instruct the model to perform tasks under different settings. In particular, we consider some very strong LLMs as our subjects, which are trained to follow human instructions. Thus, the prompt engineering approach could be used in our experiments to realize controlled experiments.

The first aspect we studied is whether LLMs align with human educators during automatic scoring. For the basic setting, we provide question descriptions to LLMs and ask them to generate analytic rubrics to grade student responses to the items. A valid analytic rubric involves multiple specific rules without overlapping. A student's response fitting more rules typically gains a higher score. If LLMs can provide rules that match the rules provided by humans, we can say that LLMs align with human educators. Our controlled experiments first examined how the baseline model performed in terms of scoring accuracy and scorning reasoning process. Further experimentations focused on three characteristics that may affect this process: (1) We first consider whether providing a \textit{human-written holistic rubric} could help LLMs understand the science item. (2) We are also curious about how \textit{analytic rubrics from other items} might help the model understand the new one. (3) We are finally interested in whether providing \textit{graded student responses} could help this process. These characteristics cover the most straightforward strategies for generating better analytic rubrics. 

The second aspect we studied was whether improving the alignment between LLMs and humans also benefits their scoring accuracy. To study this problem, we instruct LLMs to perform the traditional automatic scoring task by providing different analytic rubrics, including the human-written ones and LLM-generated ones collected from the previous experiments. \revise{In our hypothesis, if the LLM is guided by a more accurate analytic rubric (more aligned/matched with the human written ones), its automatic scoring performance will also be better.}
\revise{Here, since all assessment items are meticulously designed by human experts, we define these human-crafted analytic rubrics provided by the dataset as the ground-truth rubrics for our task. In addition, please note that a rubric leading to a better grading performance does not necessarily mean it is a correct one (please see further discussions on case studies in Section~\ref{case_study}).} 

Our experiment used the Mixtral-8x7B-instruct~\cite{jiang2024mixtral} model as our subject LLM. Mixtral is a sparse mixture-of-expert language model consisting of 47B parameters in total. As an open-sourced language model, it has surpassed those closed-source models under human evaluations~\cite{jiang2024mixtral}, such as ChatGPT-3.5 Turbo~\cite{GPT}, Claude-2.1~\cite{Claude}, Gemini Pro~\cite{Gemini}, and Llama-2 70B-Chat~\cite{Llama}. It was selected because of its superior performance and free accessibility. \revise{In addition, we follow previous work~\cite{Chain-of-thought,wu2024language} to set the hyper-parameters $temperature=0.0$ and $top-p=0.01$ during the text generation process to ensure the reproducibility of our experiments.} To better instruct Mixtral for our downstream tasks, we applied the role-play prompting strategy~\cite{zheng2024judging}. Specifically, each round of conversation with the Mixtral started with the same system prompt, i.e., ``\texttt{The assistant is an impartial science educator working in a middle school. His job is working under the supervision of the User.}'' For the analytic rubric generation task, we provide a detailed instruction as ``\texttt{Your job is to provide the analytic rubric for a science item. The analytic rubric includes a minimum set of rules, each of which covers a specific required action, and their complete collection describes the requirements of the entire task.}''
Similarly, for the automatic scoring task, Mixtral is instructed with ``\texttt{Your job is to evaluate the quality of student responses strictly following the Analytic Rubric provided previously.}''
\revise{Figure~\ref{fig:templates} provide the basic templates for both tasks.}

\subsection{Analysis Methods}
The experimental results were analyzed quantitatively. 
Specifically, to analyze the alignment gap between LLMs and humans, we measured the differences between the human-crafted and LLM-generated analytic rubrics with \textit{precision}, \textit{recall}, and \textit{F1} scores, where the precision score is the percentage of LLM-generated rules that match human-crafted ones, the recall score is the percentage of human-craft rules that have been mentioned by the LLM, and the F1 score is the geometric averaging of precision and recall scores. 
The F1 scores of certain settings are further evaluated by the Student T-Test with confidence level $\alpha=0.05$, where the null hypothesis is that the given two average scores are equal. 
\textit{Spearman's rank correlation} coefficient between the analytic rubric F1 score and automatic scoring performance was referenced to analyze the benefits of filling the alignment gap. \revise{If the Spearman's correlation is significantly greater than zero, we could say there is a correlation between the human-LLM alignment and the automatic scoring performance.}
Additionally, we conducted simple qualitative case studies on the failed analytic rules generated by LLMs to explore how different characteristics impact the alignment gap.   

\section{Experiments}
\subsection{Evaluating the alignment gap via the analytic rubric generation task.}
\vspace{0.2cm}
\subsubsection{\textbf{Settings}}
\label{evaluating}
Mixtral is instructed to generate analytic rubrics for science items from our datasets. 
For each item from the dataset, the prompt starts with the system prompt, followed by the task description about the analytic rubric generation as conversation history. By default, we provided the question, background, as well as ground truth scores for student responses to the testing item to Mixtral and requested it to provide the analytic rubric. In addition, we considered several variations, including (1) formatting the other 11 items or only one other item with human-written analytic rubrics as in-context examples (Fullshot v.s. Oneshot), (2) guiding Mixtral to provide analytic rubrics according to maximally five human-graded student responses (w/ v.s. w/o graded responses), (3) providing human-written holistic rubrics as guidance (w/ v.s. w/o holistic rubrics). 

Each generated analytic rubric includes multiple rules, which are separated by a special symbol ``$|||$'' for post-processing. 
Given a set of rules generated by Mixtral, we compared their meanings with all human written rules by using a fine-tuned RoBERTa-large~\cite{reimers2019sentence}. This fine-tuned model returns a semantic similarity score between zero and one, where returning one indicates the given pair of texts share the exact same meaning. We consider a generated rule to be correct if any human-written rule shares a 0.5 similarity with it, and then, we compute the precision score of LLM-generated rules by dividing the number of correct rules by the number of total generated rules.  Similarly, Mixtral recalls a human-written rule if it shares at least 0.6 similarities with any of the LLM-generated rules, and then we compute the recall score of human-written rules by dividing the number of recalled rules by the number of total human-written rules.  
We finally compute the F1 score as $F1=2\times (p \times r) / (p + r)$, where $p$ and $r$ refer to the precision score and recall score, respectively.
Ideally, if Mixtral understands the science items as human educators, its generated rules match all human-written rules, resulting in 1.0 precision, 1.0 recall, and 1.0 F1 scores. The average performances over 12 science items under different experimental settings are listed in Table~\ref{main_table}.

\begin{table*}[htb]
\footnotesize
\begin{center}
\caption{Performance of analytic rubric (A.R.) generation and grading tasks. \revise{The performance of the LLM-generated rubrics for the analytic rubric generation task (i.e., Precision, Recall, and F1) is evaluated against human-written rubrics.}} \label{main_table}
\begin{tabular}{l|cccc|c}
\hline
\hline
\multirow{2}{*}{\textbf{Settings}} & \multicolumn{4}{c|}{\textbf{Analytic Rubric Generation}} & \textbf{Grading} \\

& \textbf{\#Rules} & \textbf{Pre.} & \textbf{Rec.} & \textbf{F1} & \textbf{Acc. (\%)} \\

\hline
No A.R. (control) & 0.00$_{\pm 0.00}$  & 0.000$_{\pm 0.00}$ & 0.000$_{\pm 0.00}$&0.000$_{\pm 0.00}$ & 34.83$_{\pm 11.50}$\\
Human A.R. (comparison)& 2.25$_{\pm 0.83}$ &1.000$_{\pm 0.00}$ & 1.000$_{\pm 0.00}$& 1.000$_{\pm 0.00}$  & 50.41$_{\pm 12.18}$\\
\hline
One-shot A.R. &3.25$_{\pm 1.09}$ &0.525$_{\pm 0.33}$ & 0.701$_{\pm 0.34}$&0.580$_{\pm 0.32}$& 49.17$_{\pm 13.28}$\\
Full-shot A.R.  & 2.50$_{\pm 0.87}$ & 0.688$_{\pm 0.34} $ & 0.681$_{\pm 0.34}$ &0.664$_{\pm 0.33}$& 49.41$_{\pm 10.19}$\\
Full-shot A.R. + Holistic Rubrics & 2.75$_{\pm 1.16}$ &0.782$_{\pm 0.31}$ & 0.778$_{\pm 0.28}$ & 0.752$_{\pm 0.28}$ & 54.58$_{\pm 9.01}\,\,$\\
Full-shot A.R. + Graded Responses &3.33$_{\pm 1.43}$ & 0.326$_{\pm 0.33}$ & 0.528$_{\pm 0.36}$ & 0.350$_{\pm 0.28}$ & 48.41$_{\pm 10.17}$\\

\hline
\hline
\end{tabular}
\begin{tablenotes}
\footnotesize
 \textbf{A.R.:} analytic rubric. \textbf{One-shot:} randomly selects one A.R. as an in-context example. \textbf{Full-shot:} uses the rest of A.R. as in-context examples. \textbf{\#Rules:} average number of rules for each analytic rubric. \textbf{Pre.:} average precision score of generated rules. \textbf{Rec.:} average recall score of human-written rules. \textbf{Acc.:} scoring accuracy on student responses. We report the results in the format of ``avg.$_{\pm \text{std.}}$''.
\end{tablenotes}
\end{center}

\end{table*}

\subsubsection{\textbf{Results}}
\noindent\textbf{Mixtral generally understands the assessment items as human graders, but the gap still exists.} 
``Full A.R. + Holistic Rubrics'' achieves the best performance on our analytic rubric generation task with 0.782 precision, 0.778 recall, and 0.752 F1 scores. This result shows that most human-crafted analytic rubrics have been automatically developed by Mixtral as well, indicating that it understands the essential nature of assessment items generally similar to human educators. 
However, the performance standing away from 1.00 also shows that the misalignment gap between LLM and human graders exists, emphasizing the risks of replacing human graders with LLMs.

\,
\begin{figure*}[h!t]
\caption{\revise{Examples of reasons for incorrect analytic rubrics. For those incorrect rules, we underline the terms that lead to their failures. Here, the incorrect cases are fabricated for demonstration purposes.}}
\label{incorrect_examples}
\begin{tcolorbox}[colback=gray!5!white,width=1.02\textwidth,colframe=blue!75!black,]
\textbf{Ground Truth:} When the average kinetic energy of particles (or particle motion) increases, the temperature increases.\\\\
\textbf{Inappropriate Expression:} When the average kinetic energy of particles (or particle motion) \underline{decreases}, the temperature \underline{decreases}.\\\\
\textbf{Incorrect Logic Chain:} When the average kinetic energy of particles (or particle motion) increases, the temperature \underline{decreases}.\\\\
\textbf{Incorrect Logic Object:} When the average \underline{particles} increase, the temperature increases.\\\\
\textbf{No Logic Chain:} \underline{Explanation of the change in the movement of water molecules}.
\end{tcolorbox}
\end{figure*}

\noindent\textbf{Examples of analytic rubrics help Mixtral understand the assessment items closer to human graders.}
We observe that if we only present an analytic rubric from one other item as an example to Mixtral, its F1 score for the analytic rubric generation task drops from 0.664 to 0.580 compared to the ``Fullshot'' setting that provides Mixtral with analytic rubric examples from other 11 items. This means that those in-context examples of analytic rubrics from other assessment items teach Mixtral how to generate analytic rubrics that are aligned with humans. In addition, we find that the average number of rules generated under the ``Fullshot'' setting (2.50, 2.75, and 3.33) is generally closer to that of the human written rules (2.25) compared to that of ``Oneshot'' setting (3.25). This result indicates that providing example analytic rubrics from other assessment items can also improve Mixtral for the analytic rubric generation task.

\,

\noindent\textbf{Holistic rubrics help Mixtral comprehend assessment items as human graders.}
Table~\ref{main_table} also shows that the F1 score of ``Full-shot A.R. + Holistic Rubrics'' (0.752) outperforms that of ``Full-shot A.R.'' (0.664). Although the differences have not been verified by the statistic test (p-value $>$ 0.05), they still represent a notable improvement in the quality of generated analytic rubrics. 
Future work could investigate the conditions in which holistic rubrics enhance Mixtral's ability to generate higher-quality analytic rubrics.

\,

\noindent\textbf{Examples of graded responses mislead Mixtral to understand the assessment items.}
It is surprising to observe that providing human-graded student responses within the context actually leads Mixtral to generate analytic rubrics that are less similar to human graders'. In specific, the performance drops from 0.664 to 0.350. The difference is verified by statistically significant (p-value $<$ 0.05). 
This observation challenges the common sense that the human-graded student responses help LLMs to be better graders, whereas other researchers~\cite{CoT,in-context,xia2024empirical} find that incorporating these graded responses improves the automatic scoring accuracy. 
Please check Sec.~\ref{sec:case_study} for a qualitative analysis of this phenomenon. 

\subsection{Case Study on Generated Analytic Rubrics} 
\label{sec:case_study}
%\begin{wrapfigure}{r}{0.7\textwidth}
%\vspace{0.cm}
\begin{figure}
    \centering

\begin{tikzpicture}
\begin{axis}[
        width  = 0.7*\textwidth,
        height = 4.3cm,
        major x tick style = transparent,
        ybar,
        ytick={10,20,30,40,50},
        enlarge x limits=0.3,
        bar width=0.4cm,
        ylabel = {Error Causes (\%)},
        symbolic x coords={Setting 1,Setting 2},
        xtick = data,
        legend style={
      %fill,
      %at={(0.50,-0.4)},
      at={(1.35,0.8)},
      legend columns=1,
      legend cell align=left,
      anchor=north,
      },
        scaled y ticks = false,
    ]
        \addplot+[bar shift=-0.6cm] coordinates {(Setting 1,44.44) (Setting 2, 23.08) };
        \addplot+[bar shift=-0.2cm] coordinates {(Setting 1,11.11) (Setting 2, 23.08) };
        \addplot+[bar shift=0.2cm] coordinates {(Setting 1,22.22) (Setting 2, 0.1) };
        \addplot+[bar shift=0.6cm] coordinates {(Setting 1,22.22) (Setting 2, 53.85) };
        \hspace{-0.4cm}\legend{Inappropriate Expression, Incorrect Logic Chain, Incorrect Logic Object, No Logic Chain}
        \end{axis}
        \end{tikzpicture}
\caption{Causes of errors for the generated analytic rubrics. Setting 1 denotes ``One-shot A.R. Guided'', and Setting 2 refers to ``Full-shot A.R. Guided + Graded Responses''.}
\label{fig:case_study} 
\end{figure}
%\vspace{0cm}
%\end{wrapfigure}

\subsubsection{\textbf{Settings}}
This case study concentrates on offering additional insights into the observations from the previous section. Specifically, we focus on the nine assessment items where our best model (``Full-shot A.R. + Holistic Rubrics'') generates zero incorrect rules. To investigate the impacts of examples of analytic rubrics and graded responses, we perform the bad case studies on their corresponding models (i.e., ``One-shot A.R.'' and ``Full-shot A.R. + Graded Responses''). We attribute these items' incorrect rules generated by the ``Oneshot'' or ``Graded Responses'' settings to several possible causes, including ``Inappropriate expression'', ``Incorrect logic chain'', ``Incorrect logic object'', and ``No logic chain provided''. 
These causes are inspired by our expectation of LLM-generated rubrics that the valid logic chain of rules should be appropriately expressed to align with human-crafted rubrics. 
\revise{Figure~\ref{incorrect_examples} presents some fabric examples that fall into each cause for demonstration.}
The authors of this paper then annotate each incorrect rule from the generated wrong analytic rubric with one of these possible causes.  
Figure~\ref{fig:case_study} plots the proportion of errors causing generated analytic rubrics to be incorrect.

\subsubsection{\textbf{Results}}
\label{case_study}
\noindent\textbf{Examples of analytic rubrics guide Mixtral to describe rules in a way that aligns with human graders.}
Figure~\ref{fig:case_study} shows that a large proportion of causes to the incorrectly generated rubrics from the ``One-shot A.R. Guided'' setting is an inappropriate expression. Specifically, we find that some of the rules generated under the ``Oneshot'' settings were not expressed in the writing style described by human graders, resulting in the evaluation system introduced in Sec.~\ref{evaluating} treating them as incorrect ones. 
For example, while the human written one is ``\texttt{When thermal energy is transferred to the ball (outside in high temperature), air particles inside the ball move faster.}'', Mixtral under the ``Oneshot'' setting generates ``\texttt{At the lower temperature, the air molecules inside the ball move slower and occupy less space, causing the ball to appear less inflated.}''. It appears that Mixtral describes the same logic chain in the opposite direction to the human one. They are scientifically analogous as they imply symmetric phenomena from the viewpoint of professionals. Therefore, this shows that Mixtral has some scientific reasoning ability. However, it is not guaranteed whether student respondents who describe what happens at lower temperatures have a correct understanding of what will happen at higher temperatures. This shows the need for tailored scoring rubrics that specifically examine students' understanding of the problem situation, and Mixtral shows its weaknesses in this educational sensitiveness. Fortunately, further analysis shows that such inappropriate expressions can be partially overcome under the ``Fullshot'' setting.

\,

\noindent\textbf{Examples of graded responses induce Mixtral to find shortcuts for automatic scoring.}
It is obvious in Figure~\ref{fig:case_study} that most human-graded responses-guided analytic rubrics are incorrect because they fail to present a logical point for grading. In particular, we find that, by presenting examples of graded responses, Mixtral tends to provide rubrics that only summarize some superficial keywords. Take the assessment item which asks the student to ``\texttt{explain how your model shows that transferring thermal energy to water changes the movement of water molecules and temperature of water,}'' as an example, although it performs perfectly under the ``Fullshot'' and ``Oneshot'' settings, it still generates rules without any logical points after providing graded responses, such as ``\texttt{The model shows the water molecules before and after heating.}'' and ``\texttt{The model shows the kinetic energy of the water molecules before and after heating.}'' In this case, graded-responses-guided Mixtral simply considers the keywords ``water molecules'' and ``kinetic energy'' to be scoring rubrics, failing to evaluate the logic of the student responses. 
It is worth noting that if students mention these keywords in their responses, they probably have accurately described the correct logical chain. That is to say, the appearance of these keywords can be viewed as the shortcuts found by Mixtral for automatic scoring. However, these shortcuts may encourage students to take shortcuts; that is, the students may only write some keywords on their responses without presenting a valid logical chain. If the students abuse these shortcuts, the LLM graders cannot provide effective feedback to the students anymore.

\subsection{Evaluating Relations between Automatic Scoring and Analytic Rubric Generation}
\subsubsection{\textbf{Settings}}
This experiment examines the hypothesis that an LLM grader could produce more accurate scoring feedback to student responses by enhancing their understanding of the analytic rubrics. 
Specifically, we instruct Mixtral to use its generated analytic rubrics for automatic scoring. 
To do so, we design the prompt template as ``\texttt{Your job is to evaluate the quality of student responses strictly following the Analytic Rubric provided previously.}''
Ideally, higher-quality analytic rubrics could lead to better automatic scoring performance.  
Note that we do not apply any prompt engineering strategies, such as CoT and Conversational In-context Examples, to ensure that automatic scoring performance majorly relies on the quality of the provided analytic rubrics. 
For comparison, we also consider two additional controlled settings, including one without any analytic rubrics provided, as well as one having human-written analytic rubrics. 
We use \textit{accuracy} to evaluate the automatic scoring performance and report the results at Table~\ref{main_table}.

\subsubsection{\textbf{Results}}
\noindent\textbf{Higher quality analytic rubrics help Mixtral provide more accurate feedback.}
We could first observe that the Mixtral grader, without the guidance of analytic rubrics, only gets 33.5\% accuracy, indicating that it fails to provide effective feedback to students. 
On the other hand, the Mixtral grader with human-written analytic rubrics reaches 50.2\% accuracy, demonstrating a strong benefit of introducing high-quality analytic rubrics for automatic scoring. 
Meanwhile, the automatic scoring performances of Mixtral graders guided by various analytic rubrics show a strong correlation with the quality of their analytic rubrics (Spearman Rank Correlation=0.9429, p-value $<$ 0.01). 
Notably, for automatic scoring, using the analytic rubrics generated by ``Full-shot A.R. + Holistic Rubrics'' (54.58\%) outperforms using human-written rubrics (50.41\%). We suspect that this remarkable result is caused by the different writing styles between the human-crafted and Mixtral-generated analytic rubrics, where the Mixtral-generated one uses a style that could better activate Mixtral's instruction-following ability for automatic scoring. Future work could further explore how the writing style of analytic rubrics impacts the automatic scoring performance.

\section{Limitations}
This study has several limitations. The dataset may contain biases, as the assessment items were designed for specific topics in science, limiting the generalizability of findings to other disciplines or more complex tasks. Additionally, the reliance on rubric alignment as a proxy for LLM understanding assumes human-crafted rubrics are ideal, which may not always hold true. 

\section{Conclusions}
This research investigates the alignment between human and LLM graders for automatic scorings and studies the relation between the alignment and automatic scoring accuracy. 
We propose to study the alignment by incorporating the analytic rubrics of assessment items as an auxiliary subject, where the alignment gap is smaller if the LLM-generated analytic rubrics match human-crafted ones. 
%Our experiments reveal that LLM graders cannot provide analytic rubrics as human graders do, indicating their effectiveness in automatic scoring is limited by a lack of true understanding of assessment items. 
\revise{Our initial experiments reveal that LLM graders cannot provide human-like analytic rubrics, which could be the potential reason for limiting their effectiveness in automatic scoring. Further experiments show that filling this misalignment gap of LLMs could increase their automatic scoring accuracy.
Our case studies also indicate that providing some graded student responses may help LLMs find some ``tricks" for automatic scoring instead of inherently helping them better understand the assessment items.}
Our findings emphasize the potential risks of directly deploying LLMs for automatic scoring.

\revise{Future work of this study includes several directions. Firstly, it is interesting to investigate LLMs that are trained for automatic scoring and see whether their understanding of the assessment item could better align with humans. Secondly, researchers may explore other strategies to incorporate human-crafted analytic rubrics into the LLM graders, as we observed that providing human-written analytic rubrics in the prompt can increase their automatic scoring performance. Finally, researchers may study the cooperation between LLM and human graders in crafting analytic rubrics, which may significantly reduce the costs of developing new assessment items. }

\section{ACKNOWLEDGEMENT}
This work was funded by the National Science Foundation (NSF) (Award Nos. 2101104, 2138854, PI Zhai) and (Award No. 2223768, PI Liu). The findings, conclusions, or opinions herein represent the views of the authors and do not necessarily represent the views of personnel affiliated with the National Science Foundation. 

\backmatter

% \bmhead{ACKNOWLEDGEMENT}
% This work was funded by the National Science Foundation (NSF) (Award Nos. 2101104, 2138854, PI Zhai) and (Award No. 2223768, PI Liu). The findings, conclusions, or opinions herein represent the views of the authors and do not necessarily represent the views of personnel affiliated with the National Science Foundation. 

\bibliography{ours}% common bib file

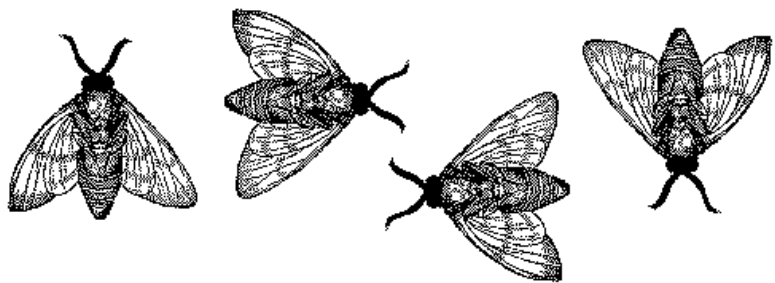
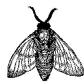
\end{document}